\documentclass{ifacconf}
\usepackage{graphicx}  
\usepackage{xcolor}
\usepackage{natbib}  
\usepackage{amsmath}
\usepackage{amssymb}
\usepackage{algorithm}	
\usepackage[noend]{algpseudocode}
\usepackage[hyphens]{url}
\usepackage{multirow}
\usepackage{array}

\renewcommand{\emptyset}{\text{\O}}             
\renewcommand{\epsilon}{\varepsilon}
\newtheorem{definition}{Definition}  

\makeatletter
\renewcommand{\nsim}{\mathrel{\mathpalette\n@sim\relax}}
\newcommand{\n@sim}[2]{%
  \ooalign{%
    $\m@th#1\sim$\cr
    \hidewidth$\m@th#1\rotatebox[origin=c]{50}{$#1-$}$\hidewidth\cr
  }%
}

\let\save@mathaccent\mathaccent
\newcommand*\if@single[3]{%
     \setbox0\hbox{${\mathaccent"0362{#1}}^H$}%
     \setbox2\hbox{${\mathaccent"0362{\kern0pt#1}}^H$}%
     \ifdim\ht0=\ht2 #3\else #2\fi
}
\newcommand*\rel@kern[1]{\kern#1\dimexpr\macc@kerna}
\newcommand*\widebar[1]{\@ifnextchar^{{\wide@bar{#1}{0}}}{\wide@bar{#1}{1}}}
\newcommand*\wide@bar[2]{\if@single{#1}{\wide@bar@{#1}{#2}{1}}{\wide@bar@{#1}{#2}{2}}}
\newcommand*\wide@bar@[3]{%
     \begingroup
     \def\mathaccent##1##2{%
          \let\mathaccent\save@mathaccent
          \if#32 \let\macc@nucleus\first@char \fi
          \setbox\z@\hbox{$\macc@style{\macc@nucleus}_{}$}%
          \setbox\tw@\hbox{$\macc@style{\macc@nucleus}{}_{}$}%
          \dimen@\wd\tw@
          \advance\dimen@-\wd\z@
          \divide\dimen@ 3
          \@tempdima\wd\tw@
          \advance\@tempdima-\scriptspace
          \divide\@tempdima 10
          \advance\dimen@-\@tempdima
          \ifdim\dimen@>\z@ \dimen@0pt\fi
          \rel@kern{0.6}\kern-\dimen@
          \if#31
               \overline{\rel@kern{-0.6}\kern\dimen@\macc@nucleus\rel@kern{0.4}\kern\dimen@}%
               \advance\dimen@0.4\dimexpr\macc@kerna
               \let\final@kern#2%
               \ifdim\dimen@<\z@ \let\final@kern1\fi
               \if\final@kern1 \kern-\dimen@\fi
          \else
               \overline{\rel@kern{-0.6}\kern\dimen@#1}%
          \fi
     }%
     \macc@depth\@ne
     \let\math@bgroup\@empty \let\math@egroup\macc@set@skewchar
     \mathsurround\z@ \frozen@everymath{\mathgroup\macc@group\relax}%
     \macc@set@skewchar\relax
     \let\mathaccentV\macc@nested@a
     \if#31
     \macc@nested@a\relax111{#1}%
     \else
          \def\gobble@till@marker##1\endmarker{}%
          \futurelet\first@char\gobble@till@marker#1\endmarker
          \ifcat\noexpand\first@char A\else
               \def\first@char{}%
          \fi
          \macc@nested@a\relax111{\first@char}%
     \fi
     \endgroup
}
\makeatother

\begin{document}
\begin{frontmatter}

\title{Efficient Computation of a Continuous Topological Model of the Configuration Space of Tethered Mobile Robots\thanksref{footnoteinfo}}
\thanks[footnoteinfo]{This publication has been supported by funding from the European Union’s Horizon 2020 Research and Innovation Programme under grant agreement No 871295 (SeaClear) and by funding from the European Union's Horizon Europe Programme under grant agreement No 101093822 (SeaClear 2.0).}

\author[First]{Gianpietro Battocletti}
\author[First]{Dimitris Boskos} 
\author[First]{Bart De Schutter}

\address[First]{Delft Center for Systems and Control, Delft University of Technology, 2628 CD Delft, The Netherlands. \\e-mails: \{g.battocletti, d.boskos, b.deschutter\}@tudelft.nl.}

\begin{abstract}
Despite the attention that the problem of path planning for tethered robots has garnered in the past few decades, the approaches proposed to solve it typically rely on a discrete representation of the configuration space and do not exploit a model that can simultaneously capture the topological information of the tether and the continuous location of the robot.
In this work, we explicitly build a topological model of the configuration space of a tethered robot starting from a polygonal representation of the workspace where the robot moves.
To do so, we first establish a link between the configuration space of the tethered robot and the universal covering space of the workspace, and then we exploit this link to develop an algorithm to compute a simplicial complex model of the configuration space. 
We show how this approach improves the performances of existing algorithms that build other types of representations of the configuration space. 
The proposed model can be computed in a fraction of the time required to build traditional homotopy-augmented graphs, and is continuous, allowing to solve the path planning task for tethered robots using a broad set of path planning algorithms.
\end{abstract}

\begin{keyword}
Tethered mobile robots, path and trajectory planning, homotopy augmentation, universal covering space, simplicial complex.
\end{keyword}

\end{frontmatter}

\section{Introduction}
\label{sec:introduction}
The structure of the configuration spaces of untethered mobile robots, and in particular their connectivity and topological properties, have been investigated extensively for their fundamental relevance in the field of motion planning \citep{latombe1991robot}.
When considering tethered mobile robots, these properties become even more relevant, as the homotopy class in which the tether lies directly impacts the ability of the robot to traverse certain paths, requiring the use of ad hoc path planning algorithms \citep{kim2014path, cao2023neptune}. 
While several approaches have been proposed to solve the problem of path planning for tethered robots, they typically rely on discrete models of the configuration space, that are not able to consider simultaneously both the topological information of the tether configuration and the continuous location of the robot \citep{igarashi2010homotopic, teshnizi2014computing, salzman2015optimal}.

In the literature, several authors propose algorithms to generate a model of the configuration space (see for example \cite{igarashi2010homotopic, kim2014path,salzman2015optimal, bhattacharya2018path}). 
However, rather than directly modeling the configuration space, those algorithms return a graph embedded in it, which presents some limits in terms of motion planning capabilities. 
The large majority of those works construct a grid graph \citep{igarashi2010homotopic, kim2014path}, which however introduce approximation errors in the modeling of the environment depending on the resolution of the grid, and can be very time-consuming to compute \citep{kim2014path}, meaning that they cannot be built online, and therefore they are not suitable for online replanning in dynamic environments.
\citet{salzman2015optimal} have proposed the use of a visibility graph instead, which, while potentially faster due to the lower number of nodes present in the graph, result in a very coarse approximation of the environment, since the nodes can only lie on the vertices of the obstacles. This can be restrictive when considering collision-avoidance guarantees and the dynamics of the robot, and may require additional steps to refine the path returned by the path planning algorithm.
An alternative approach based on a cell decomposition of the configuration space has been proposed by \citet{teshnizi2014computing}, but only for a taut tether.

In this paper we propose a novel approach to generate a complete model of the configuration space of a tethered robot, capturing information on both the robot's location and the key topological features of the tether configuration.
We do so by establishing a link between the configuration space of a tethered robot and the universal covering space of the free workspace in which the robot moves. We then exploit this result to build a computationally efficient model of the configuration space that is suitable for motion planning.
The construction of this model is performed as a preprocessing step on the free workspace, after which path planning can be performed without having to explicitly consider the shape of the tether and the homotopy class in which it lies, in contrast to other existing algorithms \citep{teshnizi2014computing, cao2023neptune}.
The proposed model yields a more complete representation of the configuration space than that of existing models, such as homotopy-augmented graphs and cell-based decompositions, has a modest computational cost to be constructed, and is continuous, allowing a flexible choice of planning algorithm (e.g., Dijkstra, A\textsuperscript{$\star$}, RRT\textsuperscript{$\star$}\ldots \citep{lavalle2006planning}), without requiring the use of algorithms specific for tethered robots.

The remainder of this paper is organized as follows. In Section \ref{sec:background} we introduce some relevant background material. In Section \ref{sec:algorithm} we define the problem setting and introduce the proposed approach to compute the simplicial complex model. 
We show some applications of this algorithm in Section \ref{sec:experiments}. 
Section \ref{sec:conclusions} concludes the paper with a summary and some future work directions.

\section{Background}
\label{sec:background}

\subsection{Homotopy classes and homotopy signatures}
Given a set $X\subseteq \mathbb{R}^2$ and two points $a, b \in X$, the set of all paths between $a$ and $b$, i.e., continuous functions $\gamma: [0,1] \rightarrow X$ with $\gamma(0)=a$ and $\gamma(1)=b$, can be partitioned in different homotopy classes, where two paths, i.e., continuous functions $\gamma_1, \gamma_2:[0,1] \rightarrow X$, with the same endpoints $\gamma_1(0) = \gamma_2(0) = a$ and $\gamma_1(1) = \gamma_2(1) = b$ belong to the same class if and only if there exists a homotopy between them, i.e., a continuous function $H:[0,1] \times [0,1] \rightarrow X$ such that $H(0, t) = a, H(1, t) = b, \forall t \in[0,1]$, and $H(s, 0) = \gamma_1(s), H(s, 1) = \gamma_2(s), \forall s \in [0,1]$ \citep{lee2010introduction}. 
We indicate homotopy equivalence between two paths as $\gamma_1\sim\gamma_2$. 
The set of all equivalence classes of loops in $X$ (i.e., paths for which $\gamma(0)=\gamma(1)$) under homotopy is known as the fundamental group and is indicated as $\pi_1(X)$.
A set $X$ is said to be simply connected if all loops are homotopic, i.e., if $\pi_1(X)$ is trivial.

One popular approach to identify the homotopy class of a path $\gamma$ is to compute its signature. Informally, a signature is a mapping $h:X^{[0,1]} \rightarrow \pi_1(X)$ that associates each path to its homotopy class. 
We indicate signature concatenation (including signature simplification)  with $\diamond$.
For more details on the construction of signatures see \citep{bhattacharya2018path}. 
Under reasonable assumptions on the workspace, signatures are homotopy invariants, i.e., $h(\gamma_1)=h(\gamma_2) \iff \gamma_1 \sim \gamma_2$ \citep[Proposition 1]{bhattacharya2018path}.

\subsection{Covering spaces}
\label{sec:covering_spaces}
Given two topological spaces $X$, $\widetilde{X}$, a covering map is a surjective function $f : \widetilde{X} \rightarrow X$ such that every point $x \in X$ has an open neighborhood $U$ whose preimage $f^{-1}(U)$ is the disjoint union of connected open subsets of $\widetilde{X}$, each of which is mapped homeomorphically onto $U$ by $f$. The space $\widetilde{X}$ is then called a covering space of $X$ \citep{lee2010introduction}. 
A covering space is called a universal covering space if it is simply connected. 
Every connected manifold has a universal covering space, which is unique up to isomorphism \citep{lee2010introduction}.

Given a covering map $f : \widetilde{X} \rightarrow X$ and a path $\gamma: [0,1] \rightarrow X$, a lift of $\gamma$ is a map $\widetilde{\gamma}:[0,1] \rightarrow \widetilde{X}$ such that $f \circ \widetilde{\gamma} = \gamma$.
Given a point $\widetilde{x} \in f^{-1}(\gamma(0))$, there exists a unique lift $\widetilde{\gamma}$ of the path $\gamma$ such that $\widetilde{\gamma}(0) = \widetilde{x}$ \citep[Corollary 11.14]{lee2010introduction}.

\subsection{Simplicial complexes and triangulations}
\label{sec:simplicial_complexes_and_triangulations}
Given a finite set of points $P \subset \mathbb{R}^2$, a geometric $k$-simplex is the convex hull of $k+1$ affinely independent points of $P$, which we denote by $s = \{p_i\}_{i=0}^k, p_i \in P$.
A 0-simplex is called a point, a 1-simplex a line segment, and a 2-simplex a triangle.
Given a $k$-simplex $s$, a (proper) face of $s$ is the convex hull of a (proper) subset of $s$. $0$-dimensional faces are called vertices, $1$-dimensional faces are called edges.
A simplicial complex $S$ is a countable set of simplices such that (i) given a simplex $s \in S$, every face of $s$ is also part of $S$, and (ii) the intersection between two simplices $s_1, s_2 \in S$ is either empty or a face of both $s_1$ and $s_2$ \citep{lee2010introduction}.
We indicate with $S_k$ the subset of $S$ of all the simplices with dimension $k$. The set $S_0$ is called the vertex set, and $S_1$ the edge set of $S$.

A triangulation of a topological space $X$ is a simplicial complex $S$ together with a homeomorphism from the underlying space of $S$ (i.e., the union of its simplices) to $X$ \citep{lee2010introduction}.
In this work we consider constrained triangulations, where the set of 0-simplices and a subset of the 1-simplices of $S$ are predefined, and the rest of the simplices are computed algorithmically to obtain a finite cover of the topological space being triangulated \citep{preparata1985computational}.

\section{Proposed algorithm}
\label{sec:algorithm}

\subsection{Problem setting}
\label{sec:problem}
We consider a bounded 2D polygonal workspace $\mathcal{W} \subset \mathbb{R}^2$ and a set of $n$ polygonal obstacles $\{O_i\}_{i=1}^n$ such that $O_i\cap O_j=\emptyset, \forall i, j \in \{1, \ldots n\}$, and we define the obstacle region as $\mathcal{O}=\cup_{i=1}^n O_i$. 
We define the free workspace as $\mathcal{W}_\mathrm{free} = \mathrm{cl}(\mathcal{W} \setminus \mathcal{O})$, which is a manifold with boundary \citep{latombe1991robot}. Without loss of generality, we assume that $\mathcal{W}_\mathrm{free}$ is connected.
Under this assumption, between any two points $a, b \in \mathcal{W}_\mathrm{free}$ there exists a shortest path which we indicate as $\widebar{\alpha}_{a, b}$ \citep{burago2001course}. We indicate the length of the shortest path between $a$ and $b$ as $\mathrm{len}(\widebar{\alpha}_{a,b})$.
Given the $n$ obstacles composing $\mathcal{O}$, we identify the $m \leq n$ obstacles that give rise to multiple homotopy classes, and for each of them we define a generator $\sigma_i$ such that (i) each generator is an infinite ray starting from a point inside the obstacle, (ii), each of the $m$ obstacles has a single generator, and (iii) the generators do not intersect with each other \citep{bhattacharya2018path}[Proposition 1]. This results in a set of generators $\{\sigma_i\}_{i=1}^m$ that can be used for the computation of homotopy signatures.

\begin{figure*}[t]
     \centering
     \includegraphics[]{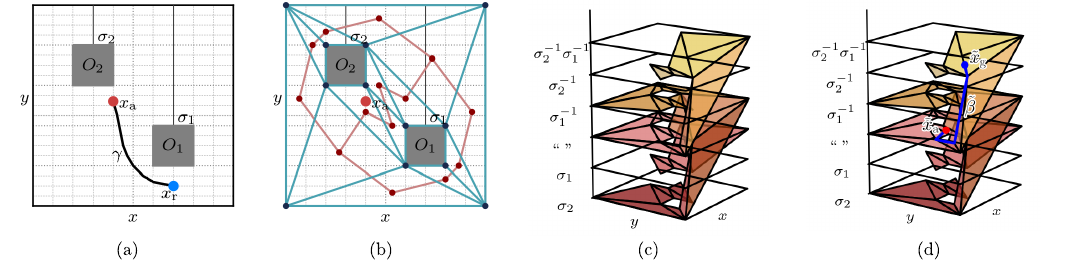}
     \caption{
          (a) Example of the workspace of a tethered robot where the robot location $x_\mathrm{r}$, the anchor point $x_\mathrm{a}$, the tether $\gamma$, the obstacles $\{O_i\}_{i=1}^n$, and the generators $\{\sigma_i\}_{i=1}^m$ are represented. 
          (b) Example of triangulation of $\mathcal{W}_\mathrm{free}$ where the primal graph $\mathcal{G}$ (blue) and the dual graph $\mathcal{G}'$ (red) are shown. 
          (c) The simplicial complex model of $\bar{\mathcal{W}}_\mathrm{free}$. On the vertical axis, the homotopy signatures corresponding to each layer are listed.
          (d) Example of path planning on $\bar{\mathcal{T}}$. To obtain an actual path for the robot to follow, it is sufficient to project $\widetilde{\beta}$ onto the base triangulation.
     }
     \label{fig:env-and-triang}
\end{figure*}

We consider a tethered robot moving in $\mathcal{W}_\mathrm{free}$ connected to an anchor point $x_\mathrm{a} \in \mathcal{W}_\mathrm{free}$ by a tether of fixed length~$l$.
\begin{definition}[Configuration of a tethered robot]
     \label{def:configuration}
     Given\\ a free workspace $\mathcal{W}_\mathrm{free}$ and an anchor point $x_\mathrm{a} \in \mathcal{W}_\mathrm{free}$, the configuration of a tethered robot is a path $\gamma$ with $\gamma(0) = x_\mathrm{a}$. 
     The configuration space is then $\mathcal{W}_\mathrm{free}^{[0,1]} = \{\gamma:[0,1]\rightarrow \mathcal{W}_\mathrm{free}, \gamma(0) = x_\mathrm{a}, \gamma \; \mathrm{continuous}\}$.
\end{definition}
The definition of configuration given above fully characterizes the configuration of both the tether and the robot in the workspace \citep{latombe1991robot, yang2022efficient_b}, since the location of the robot is $x_\mathrm{r} = \gamma(1)$.
An example of the problem setting is shown in Figure \ref{fig:env-and-triang}a.

\subsection{Reduced configuration space of a tethered robot}
The key problem of the motion planning problem for a tethered robot is the length constraint that the tether imposes, which depends on both the homotopy class where the tether lies, and on the homotopy class of the path along which the robot moves.
Between the robot location $x_\mathrm{r}$ and the goal location $x_\mathrm{g}$ there exist multiple homotopy classes, among which the planner must choose.
In fact, the fundamental group of $\mathcal{W}_\mathrm{free}$ is $\pi_1(\mathcal{W}_\mathrm{free})\cong \mathbb{F}_m$, where $\mathbb{F}_m$ indicates the free product over a set of $m$ generators \citep{bhattacharya2018path}. 
Intuitively, this represents the possibility for a loop to go around the obstacles any number of times, and in any order. 
Not all the homotopy classes are feasible, as some of them do not contain any path that can be traversed without exceeding the maximum tether length.

We next define a \emph{reduced} configuration, which only keeps track of the location of the robot and of the homotopy class of the tether instead of the full path $\gamma$.
This new configuration allows to define a one-to-one mapping between the reduced configuration space and the universal covering space $\widetilde{\mathcal{W}}_\mathrm{free}$.
The advantage of this approach is that, instead of having to keep track of the robot position \emph{and} of the homotopy class of the tether, through the reduced configuration we can map the tether configuration to a single point on $\widetilde{\mathcal{W}}_\mathrm{free}$, since the homotopy class of the tether configuration is encoded in the structure of the covering space itself. 
We define the reduced configuration as follows.
\begin{definition}[Reduced configuration of a tethered robot]
     Given the configuration of a tethered robot $\gamma \in \mathcal{W}_\mathrm{free}^{[0,1]}$ and a signature map $h$ associated to the free space $\mathcal{W}_\mathrm{free}$, the reduced configuration of the tethered robot is given by the surjective mapping $r(\gamma) = (\gamma(1), h(\gamma))$.
\end{definition}
The reduced configuration $(\gamma(1), h(\gamma))$ lives in the `smaller' configuration space $\mathcal{W}_\mathrm{free} \times \mathbb{F}_m$.
The key step is that the infinite-dimensional configuration space of paths $\mathcal{W}_\mathrm{free}^{[0,1]}$ has been replaced by the product of the 2-dimensional manifold with boundary $\mathcal{W}_\mathrm{free}$ and the countably infinite homotopy classes $\pi_1(\mathcal{W}_\mathrm{free}) \cong \mathbb{F}_m$.
In the reduced configuration we only consider information about the tether `shape' via its homotopy class, which is encoded by the signature $h(\gamma): \mathcal{W}_\mathrm{free}^{[0,1]} \rightarrow \pi_1(\mathcal{W}_\mathrm{free})$ and the endpoint of $\gamma$. 
This means that we can use the results from \citep[Theorem 11.15]{lee2010introduction} and the properties of universal covering spaces to determine a bijective mapping between reduced configurations and points in $\widetilde{\mathcal{W}}_\mathrm{free}$.
To this end, we define a mapping $g: \mathcal{W}_\mathrm{free} \times \mathbb{F}_m \rightarrow \widetilde{\mathcal{W}}_\mathrm{free}$ that connects the two spaces.
First, we select the lifted anchor point $\widetilde{x}_\mathrm{a} \in f^{-1}(x_\mathrm{a})$.
Then, for every path $\gamma$ in $\mathcal{W}_\mathrm{free}$ such that $\gamma(0)=x_\mathrm{a}$, we have a unique lift to $\widetilde{\mathcal{W}}_\mathrm{free}$.
Through this unique lift, the endpoint $\gamma_1 = x_\mathrm{r}$ of the tether is mapped to a point $\widetilde{x}_\mathrm{r} = \widetilde{\gamma}(1)$.
This lifted endpoint is shared by all paths homotopic to $\gamma$, and only by those. This results in the fact that $g$ is a bijection between the reduced configuration space $\mathcal{W}_\mathrm{free} \times \mathbb{F}_m$ and $\widetilde{\mathcal{W}}_\mathrm{free}$.
\begin{figure}[t]
     \centering
     \includegraphics{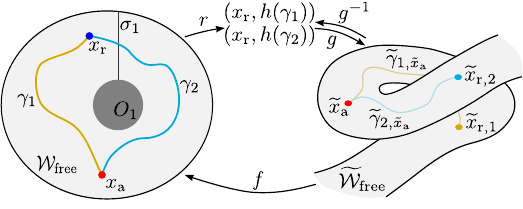}
     \caption{Example of mapping from the configurations $\gamma_1, \gamma_2$ between $x_\mathrm{a}$ and $x_\mathrm{r}$ to the reduced configurations $(x_\mathrm{r}, h(\gamma_1)), (x_\mathrm{r}, h(\gamma_2))$, and from those to the points $\widetilde{x}_{\mathrm{r}, 1}, \widetilde{x}_{\mathrm{r}, 2} \in \widetilde{\mathcal{W}}_\mathrm{free}$ through the mapping $g$.}
     \label{fig:bijective_mapping}
\end{figure}
We prove this by using \citep[Theorem 11.15]{lee2010introduction} and the fact that $\widetilde{\mathcal{W}}_\mathrm{free}$ is simply connected.
A visualization of this is shown in Figure \ref{fig:bijective_mapping}.
\begin{prop}
     \label{prop:reduced_config}
     Given a tethered robot represented by its tether configuration $\gamma$, and assuming $l=\infty$, the mapping $g: \mathcal{W}_\mathrm{free} \times \mathbb{F}_m \rightarrow \widetilde{\mathcal{W}}_\mathrm{free}$ is bijective.
\end{prop}
\begin{pf}
     Given two paths $\gamma_1, \gamma_2: [0,1] \rightarrow \mathcal{W}_\mathrm{free}, \gamma_1(0) = \gamma_2(0) = x_\mathrm{a}, \gamma_1(1) = \gamma_2(1) = x_\mathrm{r}$, their lifts to the universal covering space are unique for a fixed point $\widetilde{x}_\mathrm{a} \in f^{-1}(x_\mathrm{a})$ \citep[Corollary 11.14]{lee2010introduction}.
     If the two lifted paths $\gamma_{1, \tilde{x}_\mathrm{a}}, \gamma_{2, \tilde{x}_\mathrm{a}}$ have the same endpoint, i.e., $\gamma_{1, \tilde{x}_\mathrm{a}}(1) = \gamma_{2, \tilde{x}_\mathrm{a}}(1) = \widetilde{x}_\mathrm{r}$, then $\gamma_{1, \tilde{x}_\mathrm{a}} \sim \gamma_{2, \tilde{x}_\mathrm{a}}$ due to the simply connectedness of the universal covering space.
     This in turn means that $\gamma_1 \sim \gamma_2$ due to point (1) of \citep[Theorem 11.15]{lee2010introduction}.
     By combining this result with point (2) of \citep[Theorem 11.15]{lee2010introduction} we get that $\gamma_1 \sim \gamma_2 \iff \widetilde{\gamma}_{1, \tilde{x}_\mathrm{a}}(1) = \widetilde{\gamma}_{2, \tilde{x}_\mathrm{a}}(1)$, which means that a point $\widetilde{x} \in \widetilde{\mathcal{W}}_\mathrm{free}$ uniquely identifies a homotopy class of paths between $x_\mathrm{a}$ and $f(\widetilde{x})$ and a point in $\mathcal{W}_\mathrm{free}$, and vice versa.
\end{pf}
This result can be linked to the fact that, given a point $x \in X$, the set $f^{-1}(x)$ contains a number of copies of $x$ equal to the number of homotopy classes in which a loop passing through $x$ can lie, i.e., the number of homotopy classes in $\pi_1(X)$ \citep[Proposition 1.39]{hatcher2005algebraic}.
Therefore, in practice $g$ establishes a bijective mapping between homotopy classes of paths between $x_\mathrm{a}$ and some point $x \in \mathcal{W}_\mathrm{free}$, and the set of points $f^{-1}(x)$.

We have established a representation of a tether configuration $\gamma$ in the universal covering space $\widetilde{\mathcal{W}}_\mathrm{free}$ through the composition of the mappings $r$ (which maps $\gamma$ to the reduced configuration $r(\gamma)$) and $g$. To return to the base space, it is sufficient to apply the covering mapping $f$ to paths $\widetilde{\gamma} \in \widetilde{\mathcal{W}}_\mathrm{free}$, as represented in Figure \ref{fig:bijective_mapping}.
We highlight that the inversion of the previous composition of functions is not possible, as $r$ is not injective.

With Proposition \ref{prop:reduced_config} we have linked the reduced configuration space of a tethered robot with $\widetilde{\mathcal{W}}_\mathrm{free}$. 
However, in general the universal covering space of the free workspace of an environment with punctures (i.e., with obstacles that give rise to multiple homotopy classes) is unbounded, thus requiring some sort of truncation to make it computationally tractable. 
In case of tethered mobile robots, under the assumption of a finite tether length, a truncation criterion naturally emerges when considering only the points whose distance from $x_\mathrm{a}$ is less than or equal to the tether length $l$.
In order to do so we need to equip the universal covering space with a length structure, which assigns compatible lengths to the continuous curves on it, rendering it a metric space \citep{burago2001course}. 
To this end, we note that by the unique lifting property, once the anchor point is mapped to a fixed point $\tilde{x}_\mathrm{a} \in f^{-1}(x_\mathrm{a})$ of the covering space, we obtain a one-to-one correspondence between paths from the workspace to paths in the covering space. This yields a natural lift of the length structure from the  workspace to the covering space \citep{burago2001course}.  
Since our workspace is polygonal and admits a finite triangulation, we adopt the construction of \citep{hershberger1994computing} to build its universal covering, starting with a copy of the triangle containing  the anchor point and successively adding copies of the triangulation in a compatible manner.
As a result, lengths of paths in the universal covering space are computed by summing the Euclidean length of their restrictions to the copies of the triangulation.

This enables us to define the subset of points in $\widetilde{\mathcal{W}}_\mathrm{free}$ composed by all the points whose distance from $\tilde{x}_\mathrm{a}$ is less than or equal to $l$, i.e., the set 
\begin{equation*}
     \widebar{\mathcal{W}}_\mathrm{free}(\widetilde{x}_\mathrm{a}, l) = \{ x : x \in \widetilde{\mathcal{W}}_\mathrm{free}, \mathrm{len}(\widebar{\alpha}_{\tilde{x}_\mathrm{a}, x}) \leq l \} \subset \widetilde{\mathcal{W}}_\mathrm{free}.
\end{equation*}
We highlight that the choice of the lifted anchor point $\tilde{x}_\mathrm{a}$ is arbitrary, as the sets $\widebar{\mathcal{W}}_\mathrm{free}(\widetilde{x}_{\mathrm{a}, i}, l), \widetilde{x}_{\mathrm{a}, i} \in f^{-1}(x_\mathrm{a})$ are homeomorphic to each other.
The set $\widebar{\mathcal{W}}_\mathrm{free}$ obtained from the truncation of the universal covering space is a simply connected bounded manifold with boundary, and is topologically equivalent to the reduced configuration space of the tethered robot with finite tether length.
In the next section we propose an algorithm to compute a simplicial complex model of $\widebar{\mathcal{W}}_\mathrm{free}$.
\vfill

\subsection[Simplicial complex model]{Simplicial complex model of $\widebar{\mathcal{W}}_\mathrm{free}$}
From a computational point of view, the simplicial complex model of $\widebar{\mathcal{W}}_\mathrm{free}$ can be constructed as a homotopy-augmented triangulation. 
To do so, we start by defining the sets $V_\mathcal{O}$, $E_\mathcal{O}$ that collect the vertices and the edges\footnote{We consider all edges to be undirected, and we denote them as sets $\{a, b\}$ whose elements are the vertices connected by the edge.}  of the polygonal obstacles composing $\mathcal{O}$, respectively.
From these sets, we compute a constrained triangulation of $\mathcal{W}_\mathrm{free}$. As mentioned in Section \ref{sec:background}, the triangulation yields a simplicial complex $\mathcal{T}$ such that $\mathcal{T}_0 = V_\mathcal{O}$ and $E_\mathcal{O} \subset \mathcal{T}_1$.
The triangulation can alternatively be seen as a graph $\mathcal{G} = (\mathcal{V}, \mathcal{E})$ with $\mathcal{V} = \mathcal{T}_0$ and $\mathcal{E} = \mathcal{T}_1$.
This representation yields also the dual graph $\mathcal{G}' = (\mathcal{V}', \mathcal{E}')$, where $\mathcal{V}'$ is a collection of representative points of the 2-simplices in $\mathcal{T}_2$, and the edge set $\mathcal{E}'$ connects the representative points whose corresponding triangles are adjacent, i.e., share a 1-simplex. 
The representative points can be selected arbitrarily as long as (i) they lie in the interior of the triangles, (ii) only one representative point is selected for each triangle, and (iii) the representative points do not lie on the generators $\{\sigma_i\}_{i=1}^m$.
For convenience, in $\mathcal{V}'$ we select the representative point of the 2-simplex where $x_\mathrm{a}$ lies to be $x_\mathrm{a}$ itself.
An example of triangulation of a 2D polygonal workspace is shown in Figure \ref{fig:env-and-triang}b.
From these objects it is possible to algorithmically build the simplicial complex model of $\widebar{\mathcal{W}}_\mathrm{free}$, which we indicate as $\widebar{\mathcal{T}}$. 
The vertices $\widebar{\mathcal{T}}_0$ are tuples $(p, s), p\in \mathcal{T}_0, s \in \mathbb{F}_m$, which are mapped to $\widebar{\mathcal{W}}_\mathrm{free}$ by $g$.
Once $\widebar{\mathcal{T}}$ is computed, two homotopy-augmented graphs $\widebar{\mathcal{G}}$ and $\widebar{\mathcal{G}}'$ can be defined from $\widebar{\mathcal{T}}$ in the same way as $\mathcal{G}$ and $\mathcal{G}'$ were defined from $\mathcal{T}$.

On both simplicial complexes $\mathcal{T}$ and $\widebar{\mathcal{T}}$, distances between two points can be found efficiently using the homotopic shortest path algorithm from \citet{hershberger1994computing}. 
More precisely, to find the distance between two points $x_1, x_2 \in \widebar{\mathcal{W}}_\mathrm{free}$, it is sufficient to find a representative path $\alpha_{x_1, x_2}$ connecting them (which can be chosen arbitrarily, since $\widebar{\mathcal{W}}_\mathrm{free}$ is simply connected), find a sleeve\footnote{A \emph{sleeve} is a simply connected polygon obtained by considering a subset of a simplicial complex.} in which $\alpha_{x_1, x_2}$ lies, and apply the homotopic shortest path algorithm to find the shortest path $\widebar{\alpha}_{x_1, x_2}$ homotopic to $\alpha_{x_1, x_2}$.
This shortest path between $x_1$ and $x_2$ can be used to measure distances in $\widebar{\mathcal{W}}_\mathrm{free}$.

The algorithm to build the simplicial complex model $\widebar{\mathcal{T}}$ of $\widebar{\mathcal{W}}_\mathrm{free}$ is outlined in Algorithm \ref{alg:triangulation}. The algorithm makes use of the following functions:
\begin{itemize}
     \item $\mathtt{triangle}(v)$: given a vertex $v \in \mathcal{V}'$, returns the unique triangle $t \in \mathcal{T}_2$ where $v$ lies;
     \item $\mathtt{vertices}(t)$: given a triangle $t\in\mathcal{T}_2$, returns the set $V \subset \mathcal{V}$ of the three vertices of $t$;
     \item $\mathtt{adjacent}(v)$: Given a vertex $v \in \mathcal{V}'$, returns the set of all vertices connected to $v$ by an edge in $\mathcal{E}'$;
     \item $\mathtt{add\_simplices}(V, \widebar{\mathcal{T}})$: Given a vertex set $V$ with 3 vertices of the form $(v, s)$, with $v\in\mathcal{V}$, adds the vertices from $V$ to $\widebar{\mathcal{T}}_0$, the edges between them to $\widebar{\mathcal{T}}_1$, and the triangle they form to $\widebar{\mathcal{T}}_2$.
\end{itemize}

\begin{algorithm}
	\caption[Simplicial complex model]{Simplicial complex model of $\widebar{\mathcal{W}}_\mathrm{free}$}
	\label{alg:triangulation}
	\begin{algorithmic}[1]
		\State \textbf{Inputs}: $\mathcal{T}$, $x_\mathrm{a}$, $l$
          \State Find $\mathcal{G}$ and $\mathcal{G}'$ from $\mathcal{T}$, add $x_\mathrm{a}$ to $\mathcal{G}'$
          \State Initialize $\widebar{\mathcal{T}}\gets\emptyset$, $q_\mathrm{open} \gets \{(x_\mathrm{a}, \mathrm{``\;"})\}$, and $q_\mathrm{closed} \gets \emptyset$
          \While{$q_\mathrm{open} \neq \emptyset$} \label{line:main_loop_start}
               \State Pop $(p, s)$ from $q_\mathrm{open}$ \Comment{get point and remove it}\label{line:pop}
               \State \textbf{If} $(p, s) \in q_\mathrm{closed}$ \textbf{then} skip to next iteration
               \State $V \gets \mathtt{vertices}(\mathtt{triangle}(p))$\label{line:triangle_check_start}
               \For{$v\in V$}
                    \State $s' \gets s \diamond h(\widebar{\alpha}_{p, v})$ \Comment{signature of path to vertex}\label{line:signature_vertex}
                    \State $d \gets \mathrm{len}(\widebar{\alpha}_{x_a, v})$, with $h(\widebar{\alpha}_{x_a, v}) = s'$
                    \State \textbf{If} $d > l$ \textbf{then} skip to next iteration (line \ref{line:main_loop_start})
               \EndFor\label{line:triangle_check_end}
               \State Compute 0-simplices $(v_i, s'_i), v_i\in V, s'_i$ from line \ref{line:signature_vertex}\label{line:new_simplices} 
               \State $\mathtt{add\_simplices}(V, \widebar{\mathcal{T}})$\label{line:add_simplices}
               \For{$p' \in \mathtt{adjacent}(p)$}\label{line:adjacent_begin}
               \State $s' \gets s \diamond h(\widebar{\alpha}_{p, p'})$ \Comment{signature to adjacent node}\label{line:signature}
               \State Add $(p', s')$ to $q_\mathrm{open}$ \Comment{add new point to visit}
               \EndFor\label{line:adjacent_end}
               \State Add $(p, s)$ to $q_\mathrm{closed}$\Comment{mark point as visited}
          \EndWhile
          \State \textbf{Return} $\widebar{\mathcal{T}}$
	\end{algorithmic}
\end{algorithm}
The algorithm starts from the triangle where $x_\mathrm{a}$ lies and iteratively adds the adjacent triangles while keeping track of the homotopy class through which the representative point of each triangle is reached from.
This way, multiple copies of each triangle can be added if they are reached through different homotopy classes, corresponding to the multiple copies of $t \in \mathcal{T}_2$ in the set $f^{-1}(t) \subset \widebar{\mathcal{W}}_\mathrm{free}$. 
To this end, the dual graph $\mathcal{G}'$ is used to efficiently find adjacent triangles and to compute the signature of the path to the representative points.
After selecting a new candidate triangle $(p, s)$ in line \ref{line:pop}, lines \ref{line:triangle_check_start}--\ref{line:triangle_check_end} check whether all the vertices of the corresponding 2-simplex $\mathtt{triangle}(p)$ are reachable while respecting the tether length constraint in the appropriate homotopy class $s'$. For each vertex, $s'$ is found by concatenating the signature $s$ of the path from $x_\mathrm{a}$ to the representative point $p$ along the dual graph, and the signature of the path from the representative point $p$ to the vertex $v$ (line \ref{line:signature_vertex}).
This signature is then used in the definition of the 0-simplices $(v, s')$ used to construct $\widebar{\mathcal{T}}$ (line \ref{line:new_simplices}).
As previously mentioned, the shortest path $\widebar{\alpha}_{x_\mathrm{a}, v}$ in the homotopy class $s'$, used to check the reachability of vertex $v$, is computed using the algorithm from \cite{hershberger1994computing}. 
If all the vertices are reachable, the new simplices are added to $\widebar{\mathcal{T}}$ in line \ref{line:add_simplices}.
Finally, all the triangles adjacent to the current one are added to the open list, so that they can be checked by the algorithm at a following iteration (lines \ref{line:adjacent_begin}--\ref{line:adjacent_end}). 
An example of the algorithm's output is shown in Figure \ref{fig:env-and-triang}c.

\subsection{Motion planning on the simplicial complex model}
The simplicial complex model $\widebar{\mathcal{T}}$ can be used in different ways to perform path planning tasks. 
The primal graph $\widebar{\mathcal{G}}$ can be used with a graph search algorithm such as A$^\star$ or Dijkstra. This results in an approach equivalent to the one presented in \citep{salzman2015optimal}.
The dual graph $\widebar{\mathcal{G}}'$ can be used in the same way, with the additional advantage that, as it is a tree rooted in the triangle containing the anchor point, the search process is even faster.
In addition, differently from existing graph-based models of $\widebar{\mathcal{W}}_\mathrm{free}$, the proposed model is a simply connected manifold with a boundary, which enables the possibility to use a much broader set of path planning algorithms on $\widebar{\mathcal{T}}$ other than search-based ones, e.g., sampling-based or optimization-based ones \citep{lavalle2006planning}.
An example of path planning on $\widebar{\mathcal{T}}$ is shown in Figure \ref{fig:env-and-triang}d.

The simplicial complex $\widebar{\mathcal{T}}$ can be also used to enumerate and compare the homotopy classes through which a point $x \in \mathcal{W}_\mathrm{free}$ can be reached without violating the tether length constraint. To do so, it is sufficient to compute the preimage of $x$ under the covering map $f$, which results in the set $\{\widetilde{x}_i\}_{i=1}^k$, where $k$ is the number of sheets of $\widebar{\mathcal{W}}_\mathrm{free}$ that contain a point that maps to $x$ through $f$. 
The shortest path between these points and $x_\mathrm{a}$ can be found by applying the shortening algorithm of \citep{hershberger1994computing} to the path $\widetilde{\alpha}_{x_\mathrm{a}, \tilde{x}_i}$ (i.e., a path between $\tilde{x}_\mathrm{a}$ and $\tilde{x}_i$ in $\widebar{\mathcal{W}}_\mathrm{free}$).
This method can be used to rank all the possible homotopy classes through which a point can be reached depending on the length of the shortest path in each of them. This results in a more efficient version of the algorithm introduced in \citep{yang2022efficient}, since in our case there is no need to compute the so-called hierarchical topological tree.
An example of this ranking is shown in Figure \ref{fig:atlas-and-path-planning}, where the length of each of the 5 paths computed by the algorithm is reported below the corresponding plot. More details are provided in Section \ref{sec:experiments}.

\section{Case Study}
\label{sec:experiments}
\begin{figure*}[t]
     \centering
     \includegraphics[]{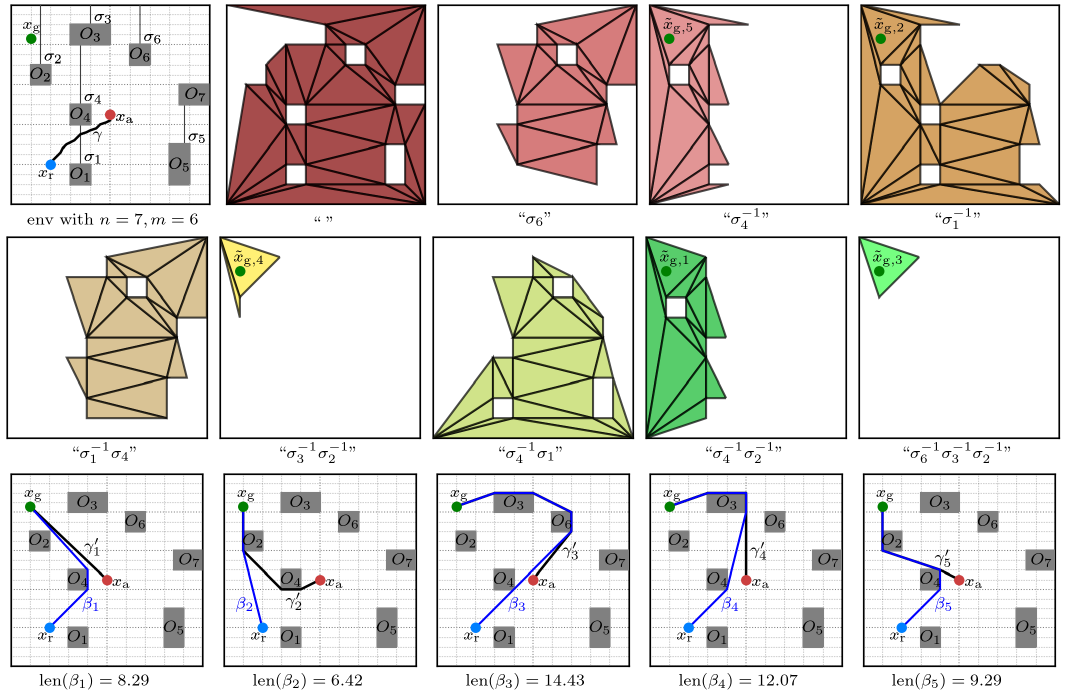}
     \caption{
          Example of path planning task. Top-left corner: initial conditions. Top two rows: The simplicial complex model $\bar{\mathcal{T}}$. For ease of visualization, the simplicial complex is displayed in a layer-by-layer fashion. Only 9 of the layers are shown in the picture for space reasons. The points $\tilde{x}_{\mathrm{g}, i}$ are indicated in the appropriate layers.
          Bottom row: The five paths $\beta_i$ computed by the path planning algorithm. The paths from $x_\mathrm{r}$ to $x_\mathrm{g}$ are displayed by the blue line, and their length is indicated below the plot. The tether configuration $\gamma_i'$ after the motion is represented by the black line. The tether is displayed as a taut line for simplicity.
     }
     \label{fig:atlas-and-path-planning}
\end{figure*}

We compare the proposed simplicial complex model with a homotopy-augmented graph model. To do so, we consider 4 environments with dimension $10\times10$ and different numbers $m \in \{1, \ldots, 8\}$ of obstacles that give rise to multiple homotopy classes, similar to that shown in Figure \ref{fig:env-and-triang}a.
For each test we specify the tether length $l$. We then generate the simplicial complex model using Algorithm \ref{alg:triangulation} and the homotopy-augmented graph using the algorithm from \citep{kim2014path}. We indicate the resulting homotopy-augmented graph as $\mathcal{H}_\mathrm{h} = (\mathcal{V}_\mathrm{h}, \mathcal{E}_\mathrm{h})$. Two versions of the graph are generated, with the grid resolution set to 0.5 and 0.25, respectively. 
The number of simplices/nodes in the resulting data structures, as well as the computation time, are reported in Table \ref{tab:comparison}.
Algorithm \ref{alg:triangulation} was implemented in Python 3.11.14 and run on a Linux server with 8 AMD EPYC 7252 (3.1 GHz) processors and 251 GB of RAM. The code is available at \url{https://github.com/gbattocletti/motion-planning-tethered-robots}.

\begin{table}[!ht]
     \caption{Comparison of $\bar{T}$ and $\mathcal{H}_\mathrm{h}$}
     \label{tab:comparison}
     \begin{tabular}{
          wc{0.12cm}wc{0.22cm}wr{0.7cm}wr{0.7cm}wr{0.85cm}wr{0.9cm}wr{0.85cm}wr{1.1cm}
          }
          \hline
          & & \multicolumn{2}{c}{$\widebar{\mathcal{T}}$} & \multicolumn{2}{c}{$\mathcal{H}_\mathrm{h}$ res. 0.5} & \multicolumn{2}{c}{$\mathcal{H}_\mathrm{h}$ res. 0.25}\\
          $m$ & $l$ & $|\widebar{\mathcal{T}}_2|$ & $t$ [s] & $|\mathcal{V}_\mathrm{h}|$ & $t$ [s] & $|\mathcal{V}_\mathrm{h}|$ & $t$ [s] \\
          \hline
          1 & 10 & 4 & 0.01 & 493 & 0.41 & 1942 & 2.50 \\
          1 & 12 & 10 & 0.01 & 679 & 0.59 & 2626 & 3.57 \\
          1 & 15 & 14 & 0.01 & 990 & 0.89 & 3845 & 6.00 \\
          1 & 20 & 26 & 0.02 & 1553 & 1.48 & 5913 & 10.26 \\
          2 & 10 & 18 & 0.02 & 679 & 0.70 & 2655 & 3.96 \\
          2 & 12 & 18 & 0.02 & 971 & 1.06 & 3885 & 6.27 \\
          2 & 15 & 42 & 0.05 & 1724 & 2.00 & 6777 & 12.81 \\
          2 & 20 & 82 & 0.11 & 3104 & 4.15 & 12039 & 32.73 \\
          6 & 10 & 157 & 0.31 & 2240 & 3.69 & 8935 & 23.87 \\
          6 & 12 & 319 & 0.68 & 4541 & 8.65 & 18175 & 68.08 \\
          6 & 15 & 995 & 2.53 & 13178 & 40.39 & 53389 & 998.65 \\
          6 & 20 & 6881 & 26.31 & 77723 & 2414.15 & 318673 & 46258.24 \\
          8 & 10 & 449 & 0.98 & 2288 & 4.08 & 8881 & 23.38 \\
          8 & 12 & 1048 & 2.71 & 5225 & 11.28 & 20347 & 95.03 \\
          8 & 15 & 4034 & 12.97 & 18176 & 69.77 & 70953 & 1984.80 \\
          8 & 20 & 39646 & 555.86 & 148246 & 9704.11 & 586482 & 171031.74 \\
          \hline
     \end{tabular}
     \vspace{0.1cm}
\end{table}

The results of the experiments reported in Table \ref{tab:comparison} show how the proposed model significantly improves on the homotopy-augmented graph model, both in terms of computation time and memory occupation, since $\widebar{\mathcal{T}}$ is formed by a much smaller number of simplices that need to be stored in memory with respect to the number of nodes in $\mathcal{H}_\mathrm{h}$. Moreover, since the resulting model is continuous, it does not introduce discretization errors, which can be costly to reduce when considering homotopy-augmented graphs, as shown by the significant increase in the graph size and computation time when using a resolution of 0.25.

Once computed, $\widebar{\mathcal{T}}$ can be used to perform motion planning tasks from the robot location $x_\mathrm{r}$ to a goal location $x_\mathrm{g}$. 
To do so, it is sufficient to (i) find the initial robot location $\widetilde{x}_\mathrm{r}$ in $\widebar{\mathcal{T}}$ through the unique lift of the tether configuration $\gamma$, (ii) find a path $\widetilde{\beta}_i$ between $\widetilde{x}_\mathrm{r}$ and one of the points in the set $f^{-1}(x_\mathrm{g})$, and (iii) project $\widetilde{\beta}_i$ to the base space $\mathcal{W}_\mathrm{free}$ to find an actual path $\beta_i$ for the robot to follow.
An example of path planning is shown in Figure \ref{fig:atlas-and-path-planning}. Given the initial tether configuration and goal location (top-left image), the simplicial complex is computed and used to solve the path planning problem. 
The number of existing solutions is equal to the number of copies in $\widebar{\mathcal{T}}_2$ of the triangle $t\in\mathcal{T}$ where $x_\mathrm{g}$ lies. In this case, 5 copies are present, corresponding to the triangle in the top-left corner of the plots representing the layers having signature ``$\sigma_4^{-1}$'', ``$\sigma_1^{-1}$'', ``$\sigma_3^{-1}\sigma_2^{-1}$'', ``$\sigma_4^{-1}\sigma_2^{-1}$'', and ``$\sigma_6^{-1}\sigma_3^{-1}\sigma_2^{-1}$''.
Each of these triangles contains a point $\widetilde{x}_{\mathrm{g}, i} \in f^{-1}(x_\mathrm{g})$, for which a path $\widetilde{\beta}_i$ can be computed in the simplicial complex. The path is then projected to the base space to obtain a path $\beta_i$ that can be traversed by the robot. The 5 paths corresponding to the 5 different solutions are represented by the blue lines in the bottom row of Figure \ref{fig:atlas-and-path-planning}. The length of each path is reported below the corresponding plot. Along with the paths, the tether configuration $\gamma_i'$ after the motion of the robot is also displayed. 
The advantage of the enumeration and comparison of all the solutions, which is enabled by the use of the simplicial complex model, can be appreciated by noting that the path $\beta_2$, which is the shortest one, results in a tether configuration $\gamma_2'$ that gets in contact with multiple obstacles and that is more twisted with respect to $\gamma_1'$, produced by $\beta_1$, despite $\beta_1$ being a longer path.

The main limitation of the proposed approach is that it is conservative when a point is close to the maximum extension of the tether. In fact, a triangle is added to $\widebar{\mathcal{T}}$ only if all its vertices can be reached within the maximum tether length. This means that, even if part of a triangle may be reachable, it is not added to the simplicial complex unless all its vertices are reachable.
This issue can be mitigated by adding a post-processing step to Algorithm \ref{alg:triangulation} that has the goal of adding additional triangles, that do not need to be copies of those in $\mathcal{T}$, and that are adjacent to the outermost ones in $\widebar{\mathcal{T}}$, making them as big as possible while respecting the maximum length constraint. This post-processing algorithm is outside the scope of this paper and will be tackled in future work.

\section{Conclusions}
\label{sec:conclusions}
We have introduced an algorithm to efficiently compute a simplicial complex model of the configuration space of a tethered robot, on which we can represent, as a single point, the location of the robot and the homotopy class where the tether lies. The model has a significantly lower computation and memory burden with respect to existing graph-based models, and is continuous, enabling the use of a broad set of algorithms to perform motion planning tasks.
We have shown the advantages of the proposed algorithm over existing approaches with comparative examples, and we have provided some examples of different motion planning operations that can be performed efficiently once the model is computed.

Future work will look at mitigating the conservativeness of the models at the limits of the configuration space through the implementation of the post-processing algorithm mentioned in Section \ref{sec:experiments}, at the extension of the new algorithm to dynamic environments, where the location of obstacles changes over time, and at the application of the proposed approach in 3D environments.

\bibliography{references}

\end{document}